%% file: root.tex
\documentclass[letterpaper, 10 pt, conference]{ieeeconf}

\IEEEoverridecommandlockouts

\usepackage{amsmath}
\usepackage{mathtools}
\usepackage{amsfonts}
\usepackage{graphicx}
\usepackage{subcaption}
\usepackage{xcolor}
\usepackage{acronym}
\usepackage{multirow}

\acrodef{6dof}[6DoF]{six degrees of freedom}
\acrodef{bow}[BoW]{Bag-of-Words}
\acrodef{slam}[SLAM]{simultaneous localization and mapping}
\acrodef{vo}[VO]{visual odometry}
\acrodef{pnp}[PnP]{Perspective-n-Points}
\acused{6dof}

\title{\LARGE \bf
	Tight Integration of Feature-based Relocalization \\in Monocular Direct Visual Odometry
}

\author{Mariia Gladkova$^{1,2}$, Rui Wang$^{1, 2}$, Niclas Zeller$^{1, 2}$, and Daniel Cremers$^{1,2}$%
	\thanks{This work was supported by the Munich Center for Machine Learning.}%
	\thanks{$^{1}$Technical University of Munich $^{2}$Artisense GmbH}%
	\thanks{Contact: {\tt mariia.gladkova@tum.de}}
}

\begin{document}
	\maketitle
	\thispagestyle{empty}
	\pagestyle{empty}

	%%%%%%%%%%%%%%%%%%%%%%%%%%%%%%%%%%%%%%%%%%%%%%%%%%%%%%%%%%%%%%%%%%%%%%%%%%%%%%%%
	\begin{abstract}
		In this paper we propose a framework for integrating map-based relocalization into online direct
		visual odometry. To achieve map-based relocalization for direct methods, we integrate image
		features into Direct Sparse Odometry (DSO) and rely on feature matching to associate online
		visual odometry (VO) with a previously built map. The integration of the relocalization poses is
		threefold. Firstly,
		they are incorporated as pose priors in the direct image alignment of the front-end tracking.
		Secondly,
		they are tightly integrated into the back-end bundle adjustment.
		Thirdly, an online fusion module is
		further proposed to combine relative VO poses and
		global relocalization poses in a pose graph to estimate keyframe-wise smooth and globally accurate
		poses. We evaluate our method on two multi-weather datasets showing the benefits of integrating
		different handcrafted and learned features and demonstrating promising improvements on camera
		tracking accuracy.

	\end{abstract}

	\keywords SLAM, relocalization, map-based localization \endkeywords

	%%%%%%%%%%%%%%%%%%%%%%%%%%%%%%%%%%%%%%%%%%%%%%%%%%%%%%%%%%%%%%%%%%%%%%%%%%%%%%%%
	\section{Introduction}
	Visual odometry (VO) and visual Simultaneous Localization and Mapping (SLAM) are important
	components of many autonomous systems that use cameras as one of their sensor modalities. For these
	systems, detection of a re-visited place can be crucial in correcting accumulated drift
	\cite{cadena2016past}, recovering from a tracking failure or solving the kidnapped robot
	problem
	\cite{fuentes2015visual}. These issues can be solved by camera-based \textit{relocalization},
	which in this work is referred to as a process of continuous online estimation of \ac{6dof}
	poses based on a
	pre-generated map. We aim to extend the conventional use of
	relocalization as a recovery module~\cite{klein2007parallel, mur2015orb} and integrate its continuous
	estimates into a \ac{vo} framework in a much more involved fashion.
	\begin{figure}[h]
		\centering
		\includegraphics[width=0.5\textwidth]{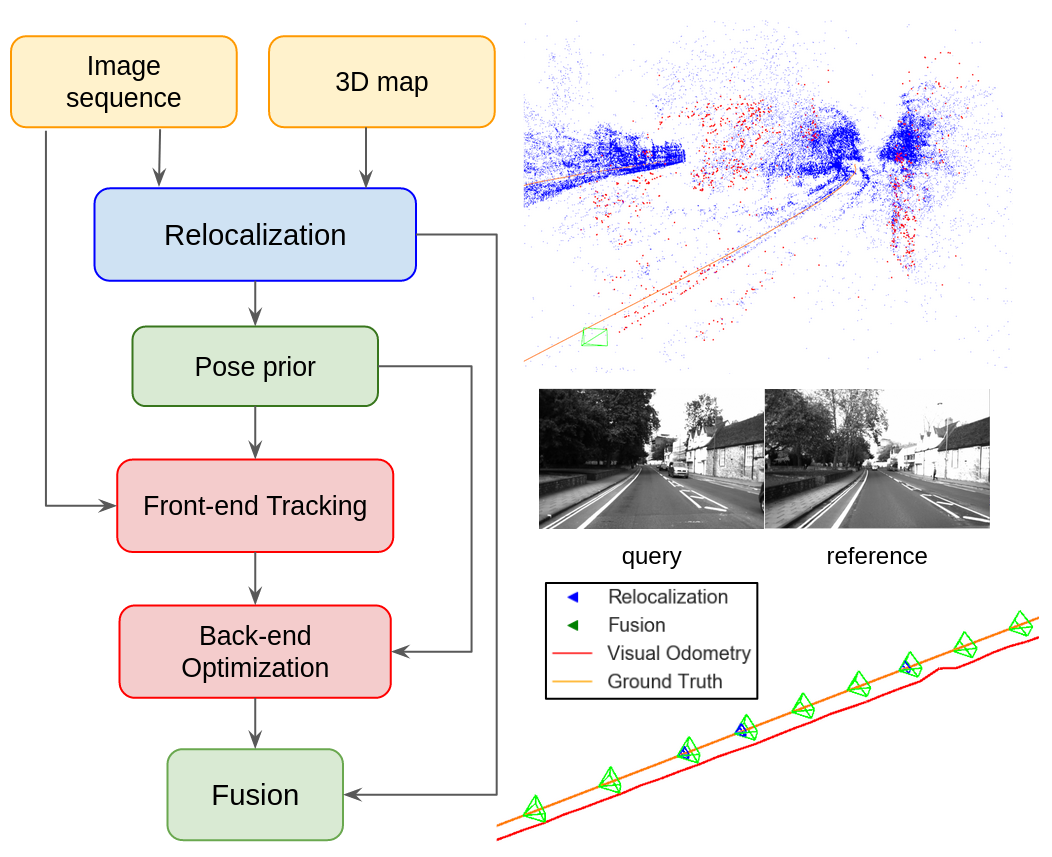}
		\caption{Left: Proposed relocalization pipeline. The relocalization module estimates global camera
			poses against a pre-built map.
			Relocalization poses are tightly integrated into the front-end tracking and the back-end
			optimization of a direct \acs{vo} framework to increase accuracy and robustness of
			camera pose estimation.
			Furthermore, the relative \acs{vo} poses and global relocalization poses are fused in a pose
			graph optimization to obtain smooth and globally accurate poses.
			Top right: Overlay of the reference map (blue) with the \acs{vo} point cloud (red).
			Bottom right: The fused poses (green)
			closely follow the ground truth trajectory (orange line).}
		\label{fig:pipeline}
		\vspace{-0.5em}
	\end{figure}

	Relocalization is a challenging task, since appearance of a map sequence can differ significantly
	from the currently acquired visual data due to weather and seasonal changes as well as human activities
	like traffic and construction.
	While feature-based VO and SLAM methods can tackle this problem by relying on the repeatability and
	descriptiveness of local features, there is no straightforward way for direct methods to achieve
	relocalization. Usually direct methods sample points that hold
	only pixel intensity values, which cannot offer any of the aforementioned feature properties.
	Only limited efforts have been made to resolve such issue. In GN-Net~\cite{von2020gn},
	the raw images are replaced by learned feature maps to enhance the invariance to brightness
	changes. LDSO~\cite{gao2018ldso}, on the other hand, proposes to integrate image features into
	DSO~\cite{engel2017direct},
	thus combining the advantages of both families. In this work, we proceed in the direction of
	merging image features into direct methods.
	When a new frame arrives, in addition to tracking features with respect to a previous
	reference frame, we also track them against a pre-built map and obtain a relocalization pose by
	feature matching. We further propose to utilize relocalization poses at three levels: by
	incorporating them as pose priors in the
	front-end tracking, by tightly integrating them into the back-end bundle adjustment (BA), and by
	fusing the global relocalization poses with relative VO estimates to get a smooth and globally
	accurate trajectory. Tight integration of pose priors into
	a direct
	sparse odometry framework is inspired by the D3VO work~\cite{yang2020d3vo}, where relative camera
	transformations estimated
	by a deep network are utilized instead. To our knowledge our work is the first approach that introduces
	online relocalization for direct VO and tightly integrates the relocalization poses into the VO optimization back-end. 
	Along with our pose graph, optimization-based fusion we
	offer a complete direct visual SLAM system that provides globally and locally accurate camera localization
	in a monocular setting. Moreover, unlike LDSO~\cite{gao2018ldso} which considers only
	ORB features~\cite{rublee2011orb} for place recognition, feature tracking and matching, we integrate
	different handcrafted and learned features to unveil their pros and cons.
	\section{Related Work}
	\subsection{Indirect versus Direct VO / SLAM}
	Indirect VO / SLAM methods~\cite{davison2007monoslam, klein2007parallel, mur2015orb} have dominated
	the field for many years. Their success can be partially attributed to robust feature detectors
	and descriptors that incorporate invariance to geometric noise, brightness and viewpoint.
	An alternative, direct formulation, which skips abstraction into a feature space and
	directly works with pixel intensities, has been firstly proposed in~\cite{jin2003semi} using an
	Extended
	Kalman Filter and was re-formulated as a non-linear optimization problem
	in~\cite{engel2014lsd,engel2017direct}.
	Direct methods sample interest points across an entire image space including edges and
	less-textured surfaces, which makes them generally more robust in cornerless environments.
	On the other hand, direct approaches are fragile to rapid motion and changes in illumination.
	Moreover, a good initialization is important to ensure optimization convergence and to guarantee an
	optimal solution. This makes direct methods inferior in wide-baseline matching, such as loop closure
	and relocalization, where global accuracy is desired.
	This issue is addressed in LDSO~\cite{gao2018ldso}, where loop closures are achieved by adapting
	a point selection strategy and by introducing local features into a direct method.
	\subsection{Handcrafted and Learned Features}
	In recent years, the number of computer vision tasks that require feature matching has
	significantly increased \cite{agarwal2011building, sattler2011fast, klein2007parallel,
		schonberger2016vote}.
	These applications introduce different feature
	requirements such as computational efficiency, invariance to scale and affine
	transformation, as well as robustness to noise and changes in lighting conditions. For many
	years
	SIFT~\cite{lowe2004distinctive} has been one of the most widely used feature descriptor, however
	its extraction is admitted to be computationally demanding~\cite{rublee2011orb}. Binary ORB
	features~\cite{rublee2011orb} that combine a FAST keypoint detector~\cite{rosten2006machine} and a
	BRIEF descriptor~\cite{calonder2010brief} have been proposed as an open-source, fast
	and lightweight alternative to SIFT. With
	the recent advances of deep learning, learned feature representations have shown a superior
	performance to handcrafted features~\cite{schonberger2017comparative}. Neural networks have been
	applied to separate tasks of keypoint localization~\cite{huang2017coarse,
		suwajanakorn2018discovery}, descriptor learning~\cite{balntas2016learning}, as well as to
	end-to-end feature extraction from images~\cite{yi2016lift, revaud2019r2d2}. In our work,
	we select three representative learned features, namely, SuperPoint~\cite{detone2018superpoint},
	R2D2~\cite{revaud2019r2d2} and ASLFeat~\cite{luo2020aslfeat}. They are integrated into a direct VO
	method and used to achieve map-based relocalization.

	\input{method}
		\begin{table*}[t!]
		\centering
		\begin{tabular}{c|c|c|c|c|c}
			\multirow{2}{*}{\textbf{Configuration}} & \multirow{2}{*}{\textbf{Odometry / Map}}
				& \multicolumn{4}{c}{no prior / prior in front-end / prior in front- and
				back-end} \\ 
			 &  & ORB & SuperPoint & ASLFeat & R2D2 \\
			\hline\hline
			same sequence
			& 03-24\_17-36-22 / 03-24\_17-36-22
			& 0.31 / 0.20 / \textbf{0.11}
			& 0.36 / \textbf{0.09} / \textbf{0.09}
			& \textbf{0.11} / 0.16 / 0.15
			& 1.40 / 0.18 / \textbf{0.17} \\

			\hline

			shadows / shadows
			& 03-24\_17-36-22 / 03-24\_17-45-37
			& 0.39 / 0.40 / \textbf{0.19}
			& 0.36 / 0.13 / \textbf{0.09}
			& \textbf{0.11} / 0.19 / 0.15
			& 1.61 / 0.20 / \textbf{0.18} \\

			sunny / sunny
			& 04-07\_10-35-45 / 04-07\_10-20-32
			& 0.42 / 0.23 / \textbf{0.19}
			& 0.48 / 0.32 / \textbf{0.17}
			& 0.49 / 0.24 / \textbf{0.15}
			& 1.22 / 0.47 / \textbf{0.42} \\

			\hline

			sunny / shadows
			&  04-07\_10-35-45 / 03-24\_17-36-22
			& 0.39 / \textbf{0.25} / 0.59
			& 0.40 / 0.32 / \textbf{0.26}
			& 0.67 / 0.41 / \textbf{0.26}
			& 1.46 / \textbf{0.71} / 0.86     \\

			\hline

			shadows / overcast
			& 03-24\_17-36-22 / 03-03\_11-52-19
			& 0.59 / 0.39 / \textbf{0.37}
			& 0.41 / 0.15 / \textbf{0.13}
			& \textbf{0.15} / 0.35 / 0.29
			& 1.35 / 0.65 / \textbf{0.53} \\

			sunny / foliage
			& 04-07\_10-35-45 / 04-23\_19-37-00
			& 0.40 / \textbf{0.35} / 0.64
			& 0.40 / \textbf{0.30} / 0.37
			& 0.69 / \textbf{0.50} / 0.54
			& 1.45 / \textbf{1.34} / 1.44 \\
			\hline
		\end{tabular}
		\caption{Relative Pose Error (RPE) on 4Seasons sequences. Each column shows the results of
			integrating different features into the direct method. The values are expressed in meters and
			computed with an interval of seven keyframes. The best results are shown in bold.}
		\label{tab:rpe_4seasons}
		\vspace{-1em}
	\end{table*}

	\begin{table*}[]
		\centering
		\begin{tabular}{c|c|c|c|c|c}
			\multirow{2}{*}{\textbf{Configuration}} & \multirow{2}{*}{\textbf{Odometry / Map}}
			& \multicolumn{4}{c}{no prior / prior in front-end / prior in front- and back-end} \\
			&  & ORB & SuperPoint & ASLFeat & R2D2 \\
			\hline \hline
			same sequence
			& 2014-12-09-13-21-02 / 2014-12-09-13-21-02
			& 0.11 / \textbf{0.10} / \textbf{0.10}
			& 0.13 / 0.11 / \textbf{0.10}
			& 0.89 / 0.26 / \textbf{0.15}
			& 0.29 / \textbf{0.10} / 0.11 \\

			\hline
			cloudy / overcast
			& 2014-11-18-13-20-12 / 2014-12-09-13-21-02
			& 0.27 / \textbf{0.22} / 0.23
			& 0.38 / 0.24 / \textbf{0.15}
			& 0.96 / 0.23 / \textbf{0.16}
			& 0.74 / 0.17 / \textbf{0.16}  \\

			cloudy / sunny
			& 2014-11-18-13-20-12 / 2015-08-12-15-04-18
			& \textbf{0.28} / 0.32 / 0.35
			& 0.35 / 0.20 / \textbf{0.16}
			& 1.08 / 0.32 / \textbf{0.17}
			& 0.73 / 0.58 / \textbf{0.49}\\

			overcast / cloudy
			& 2014-12-09-13-21-02 / 2014-11-18-13-20-12
			& 0.12 / \textbf{0.10} / 0.11
			& \textbf{0.12} / 0.14 / 0.17
			& 0.83 / 0.23 / \textbf{0.15}
			& 0.25 / \textbf{0.15} / 0.17 \\

			overcast / sunny
			& 2014-12-09-13-21-02 / 2015-08-12-15-04-18
			& \textbf{0.11} / 0.12 / 0.13
			& \textbf{0.11} / 0.14 / 0.16
			& 0.84 / 0.23 / \textbf{0.13}
			& 0.24 / \textbf{0.16} / 0.23\\
			sunny / cloudy
			& 2015-08-12-15-04-18 / 2014-11-18-13-20-12
			& \textbf{0.12} / 0.15 / 0.22
			& \textbf{0.11} / 0.12 / 0.13
			& 0.29 / 0.15 / \textbf{0.12}
			& 0.42 / \textbf{0.29} / 0.37 \\
			sunny / overcast
			& 2015-08-12-15-04-18 / 2014-12-09-13-21-02
			& \textbf{0.12} / 0.15 / 0.13
			& \textbf{0.12} / 0.14 / 0.15
			& 0.27 / 0.20 / \textbf{0.15}
			& 0.38 / 0.18 / \textbf{0.14} \\
			\hline
		\end{tabular}
		\caption{Relative Pose Error (RPE) on Oxford RobotCar sequences. Each column shows the
			results of integrating different features into the direct method. The values are expressed in
			meters and
			computed with an interval of seven keyframes. The best results are shown in bold.}
		\vspace{-1em}
		\label{tab:rpe_oxford}
	\end{table*}
	\section{Experiments}
	We choose two datasets to evaluate our method, namely the 4Seasons
	Dataset~\cite{wenzel20204seasons} and the Oxford RobotCar Dataset~\cite{maddern2017Oxford}.
	4Seasons is a novel cross-season and multi-weather outdoor dataset created by traversing nine
	different environments multiple times. It provides accurate ground truth \ac{6dof} camera poses with
	up-to centimeter precision. For our evaluations, we select one urban environment and use
	the
	sequences corresponding to six different traversals, which were captured in March and
	April of 2020. Since the sequences capture minor seasonal changes, we use them as a relatively
	less challenging setting.
	Oxford RobotCar is a large-scale dataset which is created by traversing a single route
	in Oxford for over one year. It thus contains significantly different scene layouts,
	weather and seasonal conditions. For a more challenging setting,
	we choose 3 sequences: \textit{2014-11-18-13-20} (cloudy), \textit{2014-12-09-13-21}
	(overcast) and~\textit{2015-08-12-15-04} (sunny) and use the provided Real-time Kinematic (RTK)
	poses~\cite{maddern2020real} as ground truth.
	\subsection{Integrating Pose Prior to Visual Odometry}
	To verify the benefits of integrating pose priors based on the relocalization module
	(Sec.~\ref{sec:pose_prior}) into the VO system of DSO,
	we conduct thorough experiments on the chosen datasets. For each dataset we create sequence
	pairs, such that one sequence from every pair is used for running VO, whereas the
	other is deployed for generating the map. Three settings are evaluated for each sequence pair,
	namely ``no
	prior" (i.e. conventional VO), ``prior in the front-end tracking" and ``prior in both the front-end
	tracking
	and the back-end BA". In addition, we evaluate the influence of integrating
	different feature types into the direct method, namely a handcrafted feature,
	ORB~\cite{rublee2011orb}, and
	three learned features, SuperPoint~\cite{detone2018superpoint},
	ASLFeat~\cite{luo2020aslfeat} and R2D2~\cite{revaud2019r2d2}. The relative pose error
	(RPE)~\cite{sturm2012benchmark} is adopted for quantification. As pointed out
	by~\cite{sturm2012benchmark}, rotational errors appear as translational errors when a camera
	moves, we therefore only consider the translational error in meters. The relative
	errors are computed by using an interval of seven keyframes.

	The results on the 4Seasons sequences are shown in Table~\ref{tab:rpe_4seasons}, where the rows 
	are
	grouped and
	arranged according to increasing difficulties. Note that the first row corresponds to the case
	of
	using the same sequence for the map and VO, which is idealistic and is shown as reference.
	As it can be seen from the table, the pose priors based on the relocalization poses generally
	improve camera
	tracking. Some notable exceptions
	appear with ASLFeat for the sequences with shadows, where the relocalization accuracy is not
	sufficient
	to boost pure visual odometry estimates.

\begin{figure*}[t]
	\centering
	\begin{subfigure}[b]{0.85\textwidth}
		\includegraphics[width=\textwidth]{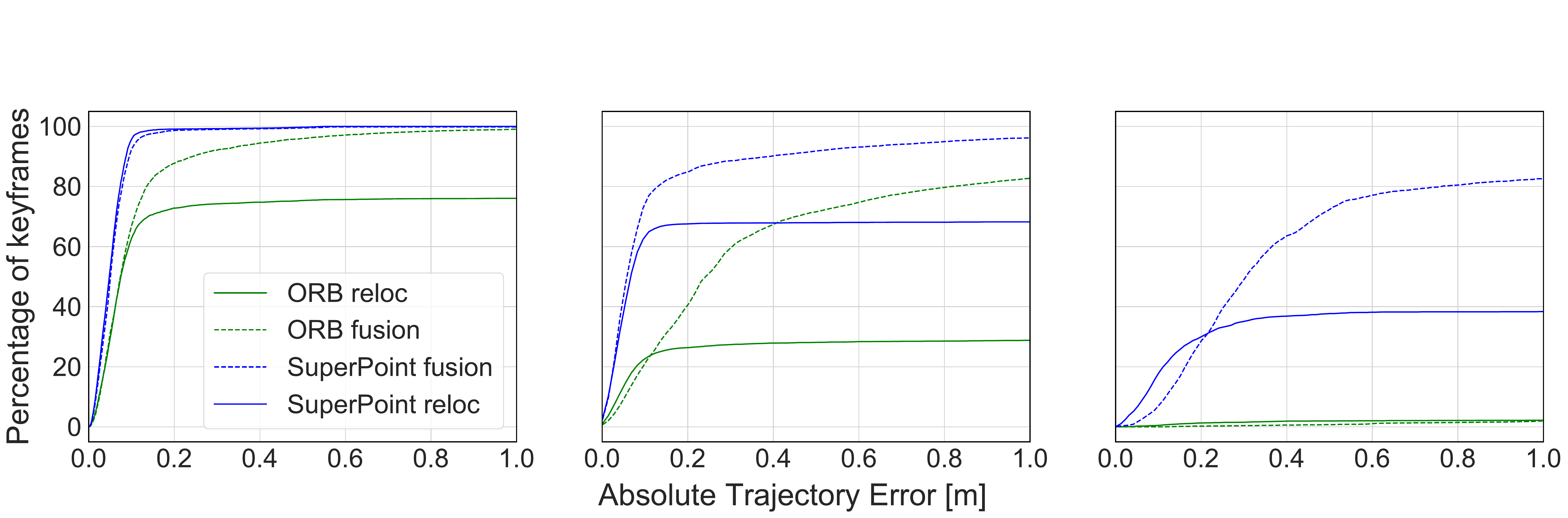}
	\end{subfigure}\\
	\hspace{1.5em}
	\begin{subfigure}[b]{0.25\textwidth}
		\includegraphics[width=\textwidth]{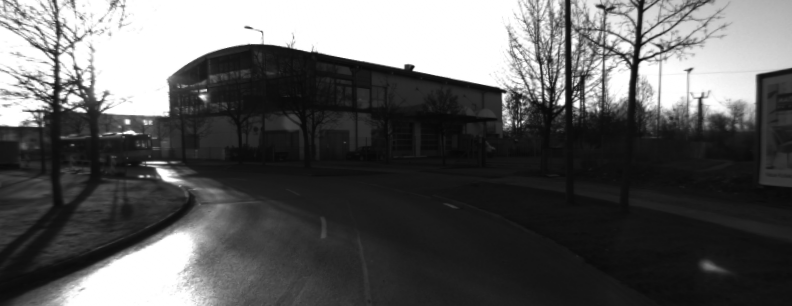}
		\includegraphics[width=\textwidth]{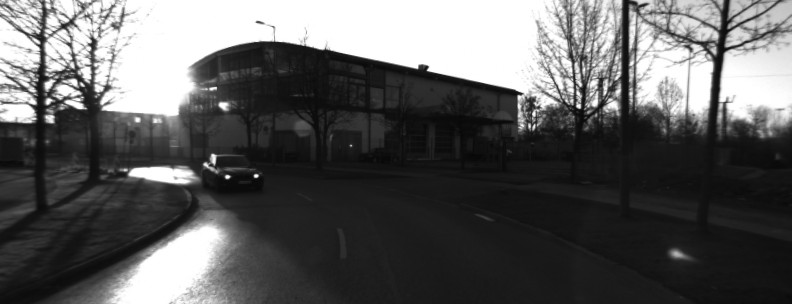}
		\caption{shadows / shadows}
	\end{subfigure}
	\hspace{2em}
	\begin{subfigure}[b]{0.25\textwidth}
		\includegraphics[width=\textwidth]{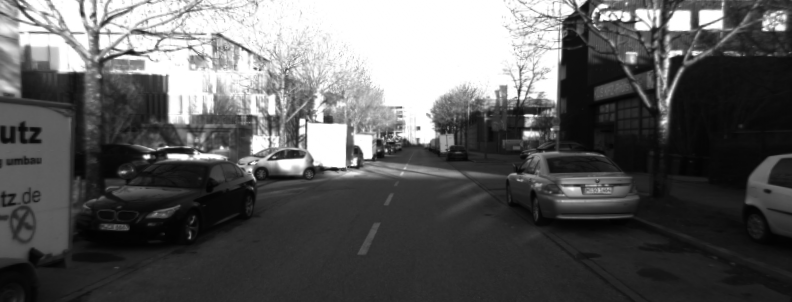}
		\includegraphics[width=\textwidth]{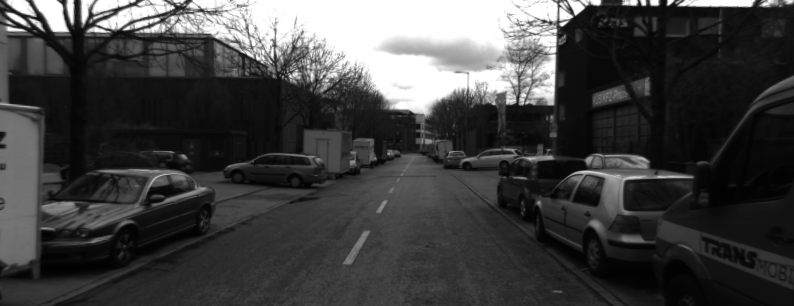}
		\caption{shadows / overcast}
	\end{subfigure}
	\hspace{2em}
	\begin{subfigure}[b]{0.25\textwidth}
		\includegraphics[width=\textwidth]{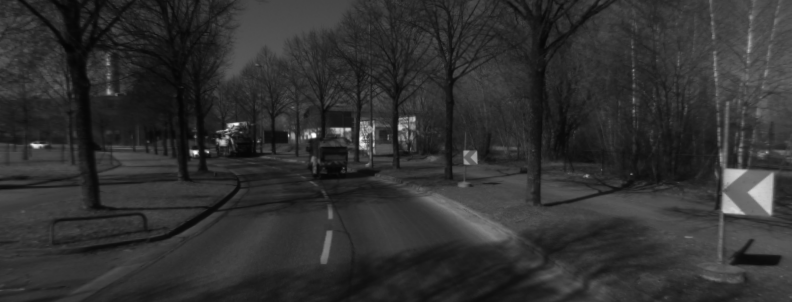}
		\includegraphics[width=\textwidth]{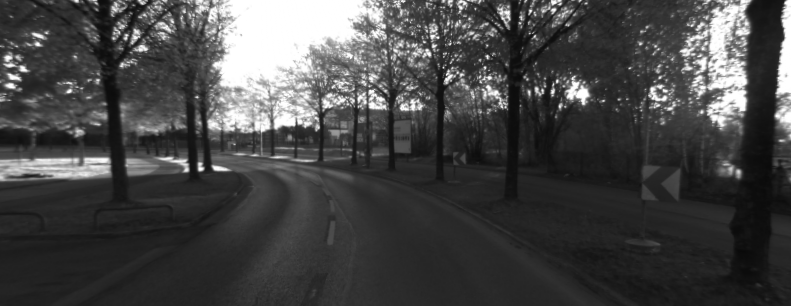}
		\caption{sunny / foliage}
	\end{subfigure}
	\caption{Cumulative Absolute Trajectory Error on 4Seasons sequences. Fusing relocalization
		poses
		with VO in a pose graph consistently improves the absolute pose accuracy. Note that
		ORB does not perform well for ``sunny / foliage'' due to the low relocalization
		success rate caused by significant appearance changes.}
	\label{fig:4Season_reloc_fusion}
	\vspace{-0.5em}
	\end{figure*}
	The results on Oxford RobotCar are presented in Table~\ref{tab:rpe_oxford}.
	For the selected pair sequences the corresponding images from the map and VO recording often
	look significantly different. Therefore, the performance of matching ORB and SuperPoint
	features starts to degrade, and the integration of pose priors does not result in the improvement when compared to pure VO.
	It should be noted, though, that despite underperforming feature matching, our integration
	maintains the system stability and does not significantly worsen the VO performance. On the other
	hand, in these more challenging conditions, relocalization based on more advanced features like
	ASLFeat and R2D2 helps to improve over pure VO, as shown in the last two columns.
	\begin{figure*}[t]
		\centering
		\begin{subfigure}[b]{0.8\textwidth}
			\includegraphics[width=\textwidth]{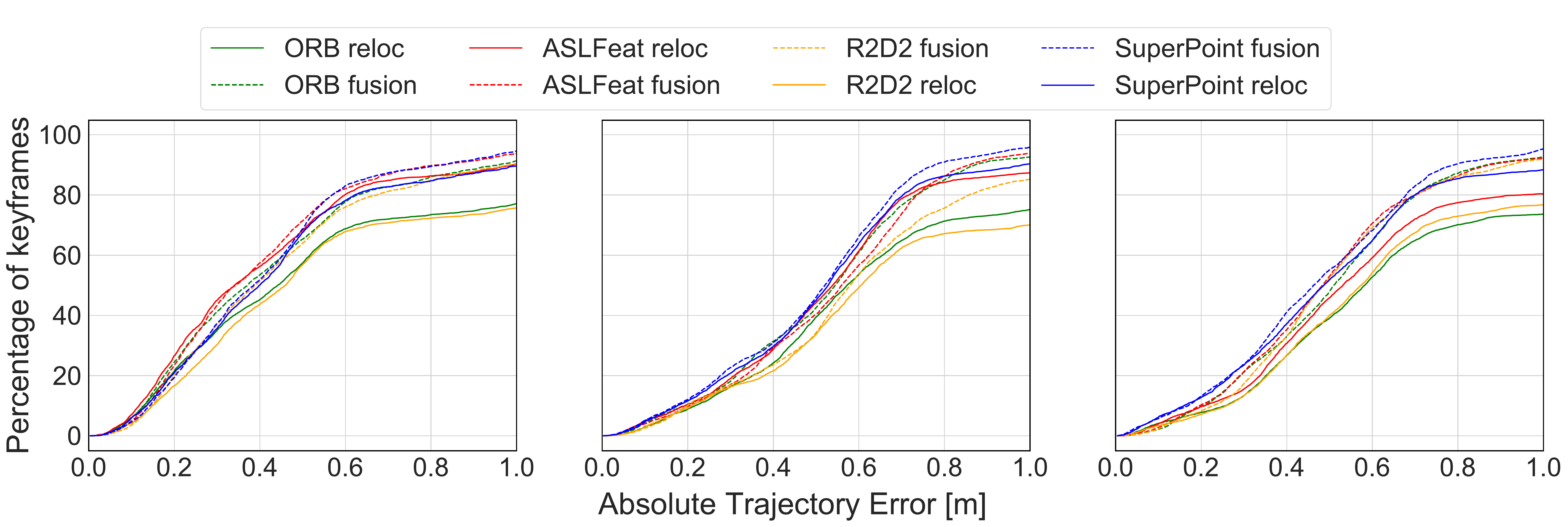}
		\end{subfigure}
		\begin{subfigure}[b]{0.27\textwidth}
			\caption{overcast / cloudy}
		\end{subfigure}
		\begin{subfigure}[b]{0.27\textwidth}
			\caption{overcast / sunny}
		\end{subfigure}
		\begin{subfigure}[b]{0.27\textwidth}
			\caption{sunny / overcast}
		\end{subfigure}
		\caption{Cumulative Absolute Trajectory Error on Oxford RobotCar sequences.
			Fusing the relocalization
			poses with VO in a pose graph consistently improves the performances.}
			\label{fig:oxford_reloc_fusion}
			\vspace{-1em}
		\end{figure*}
	\subsection{Map-Based Relocalization and Fusion with VO}
	In this section, we verify the quality of the global relocalization poses and the benefit of
	fusing them
	with the VO output. As explained in Section~\ref{sec:fusion},
	our fusion method can estimate global poses defined in the reference
	coordinate system of the map. This makes it possible to evaluate the global pose errors. In all
	the following experiments, the absolute trajectory error (ATE)~\cite{sturm2012benchmark} is
	used.

	On the 4Seasons dataset we select three sequence pairs with increasingly challenging 
	configurations on
	weather
	and
	seasonal conditions, namely ``shadows / shadows'', ``shadows / overcast'' and ``sunny / foliage''.
	The two
	features
	that work dedicatedly on grayscale images, namely ORB and SuperPoint, are evaluated. The
	cumulative
	error plots together with some example images from the odometry and the map sequences are shown in
	Fig.~\ref{fig:4Season_reloc_fusion}. It is apparent that fusing the relocalization poses with the
	VO results consistently improves pose accuracy. It is worth noting that the relocalization curves
	often saturate to values less than 100$\%$, which means we do not get relocalization poses for
	all keyframes. Yet our fusion unquestionably boosts the performances in those cases. Due to the
	significant differences caused by seasonal change, relocalization based on ORB features is
	unsuccessful for the majority of keyframes in the configuration of ``sunny / foliage''. Therefore,
	fusion estimates are not globally accurate in this case.

	We further verify our relocalization and fusion on the Oxford RobotCar dataset, using the same
	sequence
	pairs
	as in the previous section. All the four selected features are tested and the cumulative absolute errors are shown in Fig.~\ref{fig:oxford_reloc_fusion}. Despite the more
	challenging configurations
	compared to the 4Seasons experiments, our fusion consistently improves the performances over
	relocalization for all the tested features.
	\subsection{Runtime}
	Table~\ref{tab:runtime} presents a timing assessment of our system. The values are collected on
	a machine with Intel Core i7-8700K CPU, 32 GB RAM. We demonstrate
	results for ORB and SuperPoint features to account for the differences in their extraction and
	description.
	\begin{table}
	\centering
	\begin{tabular}{c|cc|c|c|c}
		\multicolumn{3}{c|}{\multirow{2}{*}{\textbf{Task}}} & \multirow{2}{*}{\textbf{Rate}}
		& \multicolumn{2}{c}{Mean $\pm$
			std {[}ms{]}} \\
		\multicolumn{3}{c|}{} &   & ORB & SuperPoint \\ \hline \hline
		\multirow{4}{*}{\rotatebox[origin=c]{90}{Reloc}} &
		\multicolumn{2}{c|}{Feature Extraction}   & F
		& 5.63 $\pm$ 1.04     & 220.93 $\pm$ 5.70    \\
		& \multirow{2}{*}{IR}     & BoW             & \multirow{2}{*}{F} & 2.85 $\pm$ 0.27
		& 2.26 $\pm$ 0.33       \\
		&                         & Spatial BoW     &                    & 7.08 $\pm$ 2.79
		& 8.50
		$\pm$ 4.51      \\
		& \multicolumn{2}{c|}{Pose Estimation}      & F                  & 8.93 $\pm$
		7.27      & 9.60
		$\pm$ 4.16       \\ \hline
		& \multicolumn{2}{c|}{Coarse Tracker}       & F                  &
		\multicolumn{2}{c}{2.66 $\pm$
			13.06}         \\
		& \multicolumn{2}{c|}{BA + Marginalization} & KF                 &
		\multicolumn{2}{c}{60.60
			$\pm$ 11.93}      \\
		& \multicolumn{2}{c|}{Fusion}               & KF                 &
		\multicolumn{2}{c}{3.54 $\pm$
			1.11}         \\ \hline
	\end{tabular}
	\caption{Runtime evaluation in milliseconds on the 4Seasons dataset.
		Relocalization is performed on the shadows / shadows odometry-map setting.
		Tasks run at the frame (F) or the keyframe-rate (KF) (5 - 10 keyframes per second~\cite{engel2017direct}).}
	\vspace{-2em}
	\label{tab:runtime}
	\end{table}
	
	As it can be seen from the table, relocalization based on ORB features is fast, which allows
	tracking to be performed in real-time. Inference of SuperPoint features remains a bottleneck, which
	can be mitigated by porting inference to GPU and incorporating further optimization techniques.
	To
	adhere to spatial BoW image retrieval, multi-level histograms can be computed in parallel.
	BA and fusion run at a keyframe rate and, therefore, are not limiting the real-time performance.
	\section{Conclusion}
	In this paper we present a complete framework which combines direct \ac{vo} and feature-based
	relocalization in an online and tightly-coupled fashion. We extensively evaluate our approach on two
	multi-weather datasets.
	Our experiments show that by integrating pose priors obtained from relocalization into both the
	front-end tracking and the back-end optimization of a direct VO method, we can significantly
	improve the tracking accuracy.
	We also show that the proposed fusion module is able to estimate globally accurate poses, even when
	relocalization is not successful for every frame.
	Furthermore, using our pipeline we leverage the strengths and
	uncover some of the weaknesses of different feature types.
	We hope that our work has revealed the power of combining
	direct and indirect approaches in the context of \ac{slam} and that it will drive further research
	in this direction.

	%%%%%%%%%%%%%%%%%%%%%%%%%%%%%%%%%%%%%%%%%%%%%%%%%%%%%%%%%%%%%%%%%%%%%%%%%%%%%%%%

	\bibliographystyle{ieeetr}
	\bibliography{references}
\end{document}

%% file: method.tex
\section{System Overview}
In the following sections we will describe in detail the proposed SLAM and relocalization framework as shown
in Fig.~\ref{fig:pipeline}.
Our pipeline consists of three major modules: 1) a relocalization module that obtains
reference poses with respect to a pre-build map (Sec.~\ref{sec:relocalization}); 2) a VO module
that integrates the relocalization information to perform robust and accurate camera tracking
within a local coordinate frame (Sec.~\ref{sec:visual_odometry}); 3) a fusion module that fuses
global map-based relocalization poses and relative visual odometry poses to obtain a smooth and
globally accurate camera trajectory (Sec.~\ref{sec:fusion}).
While our VO module uses information from the relocalization module, it is also used to generate the
map we localize against.
We will first describe our VO approach and afterwards proceed with the relocalization
module. Finally, we will explain how both components are integrated in the fusion module.

\section{Visual Odometry}\label{sec:visual_odometry}
Our VO module builds on top of DSO~\cite{engel2017direct}, a state-of-the-art
direct visual odometry algorithm.
For each new frame DSO estimates its initial pose with respect to a reference
keyframe by direct image alignment.
Poses of keyframes are then refined in a sliding window, where bundle adjustment jointly optimizes
the depth of points and all keyframe poses by minimizing a photometric energy function
\begin{equation}
\label{eq:photo_energy}
E_\text{photo} = \sum_{i \in \mathcal{F}} \sum_{\mathbf{p} \in \mathcal{P}_i} \sum_{j \in
	obs(\mathbf{p})} E_{\mathbf{p}j},
\end{equation}
where $\mathcal{F}$ is a set of all keyframes, $\mathcal{P}_i$ a set of points hosted in a
keyframe $i$, and $obs(\mathbf{p})$ a set of keyframes that observe a point $\mathbf{p}$.
$E_{\mathbf{p}j}$
is a weighted photometric error term for a point $\mathbf{p}$ hosted in a frame $i$ and observed in
a frame $j$.
For details on the energy formulation please refer to~\cite{engel2017direct}.

\subsection{Pose Priors}\label{sec:pose_prior}
To improve the accuracy and robustness of \ac{vo}, we use the information gained
from relocalization against a pre-built map (Sec.~\ref{sec:relocalization}). The relocalization
poses are used as priors for both the front-end tracking (coarse tracker) and the back-end bundle
adjustment.

\subsubsection{Pose Prior for Coarse Tracker}\label{sec:coarse_tracker_prior}
In the tracking front-end we use a relative pose prior based on the global poses obtained
from the relocalization module (Sec.~\ref{sec:relocalization}).
This prior serves as initialization for the two-frame direct image alignment and replaces the motion model
analogue. In addition, we construct a factor
graph corresponding to the coarse-to-fine pose refinement and introduce the prior based on the relocalization poses as a factor between
a reference keyframe and the current frame.
If the pose prior is unavailable due to an unsuccessful relocalization attempt, the front-end is initialized according to a
constant motion model, as described in~\cite{engel2017direct}.

\subsubsection{Pose Prior for Bundle Adjustment}\label{sec:bundle_adjustment_prior}
Keyframe poses optimized in the bundle adjustment are defined with respect to a common local
coordinate frame.
\begin{figure}[t]
	\centering
	\includegraphics[width=0.38\textwidth]{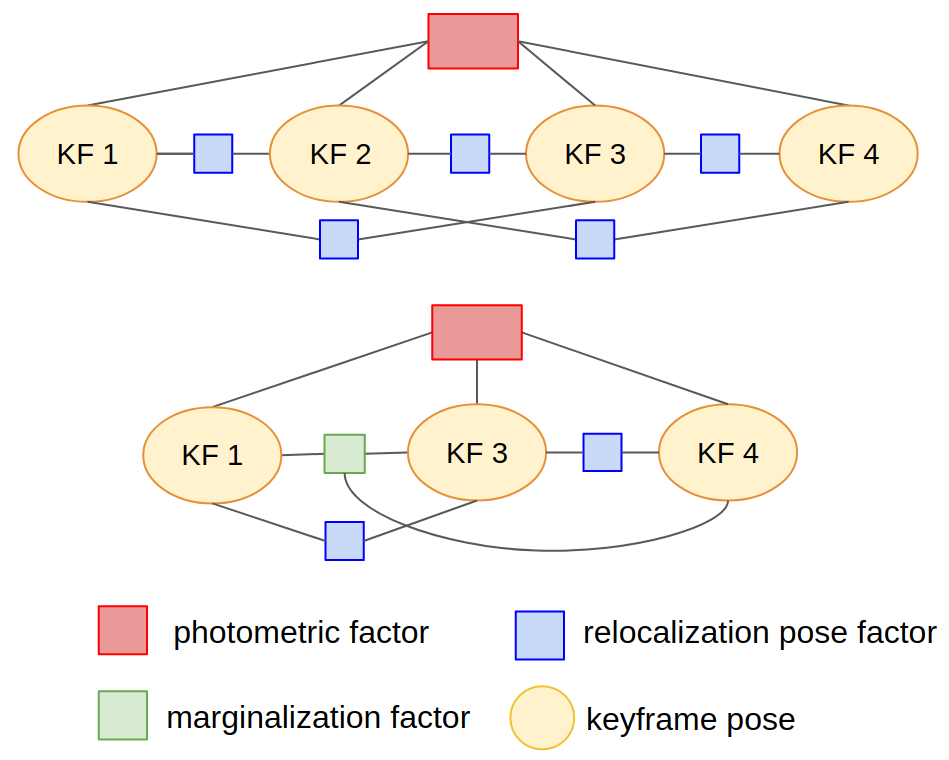}
	\caption{Example of a factor graph with 4 keyframes in the back-end
		optimization. To emphasize our contribution, some variables such as observed points are not shown (see Fig. 5 in~\cite{engel2017direct}). Each
		relocalization factor represents a relative relocalization pose between corresponding
		keyframes. When a keyframe is marginalized (bottom), residual energy after the Schur complement is
		kept as a factor (green).}
	\vspace{-1em}
	\label{fig:factor_graph_ba}
\end{figure}
Therefore, one could think of lifting this common coordinate frame to a global frame based on the
information obtained from relocalization and performing
optimization of the global poses directly in the bundle adjustment.
However, due to the marginalization of keyframes, this leads to numerical instabilities, especially
in situations when relocalization poses are sparse or not available for the first frame.
Hence, similar to the coarse tracker prior, we utilize relative pose priors between the keyframes
respectively. We can derive a factor graph as shown in
Fig.~\ref{fig:factor_graph_ba}, where the red boxes refer to the photometric factors defined in
Eq.~\eqref{eq:photo_energy} and the green boxes refer to the marginalization factors.
Relocalization factors (blue boxes) are imposed in the form of priors  $\widehat{\mathbf{T}}^j_{i}$
on
the relative pose $\mathbf{T}^j_{i} \in$ SE(3) between keyframes $i$ and $j$ according to
Eq.~\eqref{eq:energy_pose}.
\begin{equation}
\label{eq:energy_pose}
E_\text{pose} = \sum_{i \in \mathcal{F}} \sum_{\substack{j \in \mathcal{R}_i\\ j < i}}
\mbox{Log}(\widehat{\mathbf{T}}^i_{j}\mathbf{T}^{j}_{i})^T\mathbf{\Sigma}^{-1}\mbox{Log}(\widehat{\mathbf{T}}^i_{j}\mathbf{T}^{j}_{i})
\end{equation}
Here $\mathcal{F}$ is a set of all keyframes and
$\mathcal{R}_i$ is a subset of $\mathcal{F}$, which includes keyframes that have a
relocalization pose.
In our work we limit $|\mathcal{R}_i| \leq 2$.
When selecting keyframes for $\mathcal{R}_i$, the priority is given to the later ones, since the
oldest keyframes can be shortly scheduled for marginalization.
The inverse  of a covariance matrix
$\mathbf{\Sigma}^{-1} \in {\rm I\!R}^{6 \times 6}$ is modeled as
a constant diagonal matrix and Log$(\cdot)$
is a mapping from an element of the Lie group SE(3) to its twist coordinates in $\mathfrak{se}(3)$.

Combining photometric and relocalization factors, the total objective function becomes
\begin{equation}
E_\text{total} = E_\text{photo} + wE_\text{pose},
\end{equation}
where $w = 10^{3}$ is introduced to mitigate a large photometric error.
$E_\text{photo}$ and $E_\text{pose}$ are defined as in Eq.~\eqref{eq:photo_energy} and
Eq.~\eqref{eq:energy_pose} respectively.
The minimization of $E_\text{total}$ is performed in a Gauss-Newton optimization scheme.
\subsection{Feature Tracking}
While photometric formulations show superior performance with respect to \ac{vo}, they struggle in
tasks like loop closure and relocalization, since in these cases a good initialization and
photometric consistency cannot be guaranteed.
To be able to solve these problems, we follow the idea of LDSO~\cite{gao2018ldso}, which
replaces a subset of the tracked and optimized points by keypoints with associated local descriptors.
Since keypoints are tightly integrated into the photometric bundle adjustment, their accurate
depth is estimated using an entire optimization window.
While LDSO limits the use to handcrafted ORB features~\cite{rublee2011orb}, we keep our pipeline
more general,
which enables integration of any local keypoint descriptors, including learned ones.

The tracked features now can be used to solve tasks like loop closure to generate globally
consistent maps or to perform relocalization against a pre-build map
(Sec.~\ref{sec:relocalization}).

\section{Relocalization}\label{sec:relocalization}
Relocalization is carried out in a two-step approach.
Firstly, we find potential candidates in the map database using \ac{bow} image retrieval
(Sec.~\ref{sec:bow_image_retrieval}).
Secondly, a relative pose between a current frame and its map-based reference is estimated from
feature correspondences and a global relocalization pose is computed
(Sec.~\ref{sec:pose_refinement}).

\subsection{\acl{bow} Image Retrieval}\label{sec:bow_image_retrieval}

After the system has received a new image, it extracts local 2D features
and converts them to a global descriptor using a BoW database.\footnote{We use fbow, a fast version of
DBoW2/DBoW3 libraries \cite{galvez2012BoW}.} Since such a
representation does not preserve the order of features in the image, it removes the spatial
information of the feature layout and offers only a limited
description capability. To circumvent this problem, we follow the \textit{pyramid matching} method
proposed in \cite{grauman2005pyramid}. In particular, we switch to a multi-level
representation of an image, which can be intuitively viewed as placing a grid of increasingly
coarser resolution and aggregating the features in each grid cell for local histogram computations.
We
refer to Eq.~(3) of \cite{lazebnik2006beyond} for further details of the underlying approach.

To limit the number of images considered
for the similarity measure computation, we take advantage of the sequential nature of our queries and
assume that the correct tracking references lie spatially close for consecutive frames.

\subsection{Pose Refinement}\label{sec:pose_refinement}
We select the top three retrieval candidates from a pre-built map and proceed to feature
matching. False correspondences are pruned using Lowe's ratio test \cite{lowe2004distinctive} with
threshold $\tau = 0.85$. Having 3D - 2D correspondences between a reference map frame and a current
frame
we can estimate \ac{6dof} relative transformation
using a \ac{pnp} algorithm in a RANSAC scheme~\cite{fischler1981random} and refine it by minimizing
a
geometric
projection error. The final global pose is computed by
concatenating the
respective relative transformation to the global pose of a map candidate that has the biggest number
of feature correspondences.
\begin{figure}[t]
	\centering
	\includegraphics[width=0.5\textwidth]{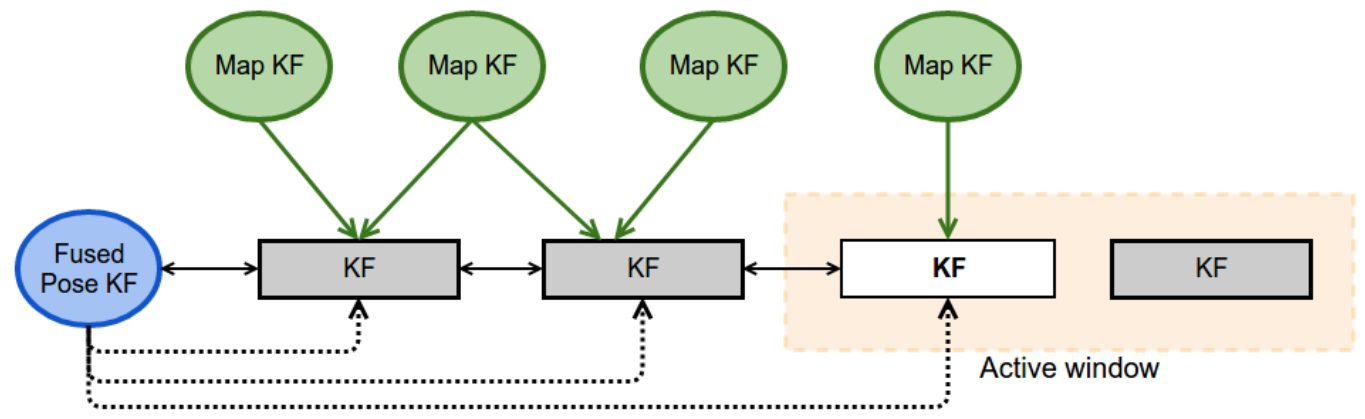}
	\caption{Example of a fusion pose graph. Circular components represent fixed variables. Solid arrows refer to imposed constraints on the optimized variables,
	whereas dotted arrows represent their initialization. The direction of arrows
	depicts relative transformation between corresponding coordinate systems. A white
	rectangle denotes a keyframe that is scheduled for marginalization.}
	\vspace{-1em}
	\label{fig:fusion_graph}
\end{figure}
\section{Fusion}\label{sec:fusion}
In addition to the integration of relocalization poses into direct image alignment and bundle
adjustment, we propose a local pose graph to fuse odometry and
relocalization estimations online. Since our relocalization
module computes relative poses with respect to global map poses, they are
suitable as pair-wise pose observations in a traditional pose graph framework. To avoid
inconsistencies caused by updates in the active window, we base our pose graph on marginalized
keyframes.

The main objective of the local pose graph optimization lies in the estimation of a fused pose
$\mathbf{F}_m \in$
Sim(3) for a
keyframe $m$ that has been scheduled for marginalization. To build the graph we consider
only keyframes that appear earlier in the sequence and have a relocalization pose. After all
keyframes are chosen, pose values are
initialized based on the oldest keyframe, which has
already received a fused pose. Specifically, the initialization of estimated fused pose
$\mathbf{F}_j$ of a keyframe $j$ is achieved by concatenating
a relative keyframe pose $\mathbf{T}^j_i$ to a fused pose $\mathbf{F}_i$ of a reference keyframe $i$, i.e.
$\mathbf{F}_j \coloneqq \mathbf{T}^j_i \mathbf{F}_i$. An example of the proposed pose graph can be seen in Fig.~\ref{fig:fusion_graph}.

We distinguish two types of constraints, namely odometry-based $e_{i,j}$ and map-based $l_{i,k}$,
which are defined in Eq.~\eqref{eq:odom_constraint} and Eq.~\eqref{eq:map_constraint} respectively. In Eq.~\eqref{eq:map_constraint} $\mathbf{M}$ denotes a map trajectory and $\mathbf{\widehat{T}}$ corresponds to global relocalization poses.
\begin{equation}
\label{eq:odom_constraint}
e_{i,j} \coloneqq \mbox{Log}(\mathbf{T}^i_{j}\mathbf{F}^{j}_{i})
\end{equation}
\begin{equation}
\label{eq:map_constraint}
l_{i,k} \coloneqq
\mbox{Log}((\mathbf{\widehat{T}}_{i}\mathbf{M}^{-1}_{k})^{-1}\mathbf{F}_{i}\mathbf{M}^{-1}_{k})
\end{equation}
In Eqs.~\eqref{eq:odom_constraint} and \eqref{eq:map_constraint}, Log$(\cdot)$ defines the mapping 
from an element of the Lie group Sim(3) to its tangent space coordinates in $\mathfrak{sim}(3)$.
The total energy, which is minimized in the local pose graph with $N$ keyframes, is represented by
Eq.~\eqref{eq:pose_graph_eq}:
\begin{align}
\label{eq:pose_graph_eq}
E_\text{fusion}&= E_\text{vo} + wE_\text{map}\\
&= \sum_{i, j \in \mathcal{F}_m} e^T_{i,j} \mathbf{\Sigma}^{-1} e_{i,j} + w \cdot \sum_{i \in
	\mathcal{F}_m} \sum_{k \in \mathcal{L}_i} l^{T}_{i,k}\mathbf{\Lambda}^{-1}l_{i,k}, \nonumber
\end{align}
where $\mathcal{F}_m$ is a set of selected keyframes together with the keyframe
$m$, $\mathcal{L}_i$ is a set of tracking references for the keyframe $i$. Lastly,
$\mathbf{\Sigma}^{-1}, \mathbf{\Lambda}^{-1} \in {\rm I\!R}^{7 \times 7}$ are inverses of
covariance matrices, which are modeled as constant diagonal matrices. Since the relocalization poses are
computed using the PnP algorithm and 3D-2D feature correspondences (Sec.~\ref{sec:relocalization}), an
odometry scale is not recovered. Thus, we set a corresponding entry in the covariance matrix $\boldsymbol{\Lambda}$
to a large number. A weighting factor $w = 10^{2}$ is chosen empirically.

In our implementation, we take advantage of the possibility of having several relocalization references
per keyframe and impose at most two measurement constraints from the map. For optimizations we fix
all map poses together with the reference keyframe.
Our pose graph optimization is implemented based on g2o~\cite{grisetti2011g2o}.

%% file: root.bbl
\begin{thebibliography}{10}

\bibitem{cadena2016past}
C.~Cadena, L.~Carlone, H.~Carrillo, Y.~Latif, D.~Scaramuzza, J.~Neira, I.~Reid,
  and J.~J. Leonard, ``Past, present, and future of simultaneous localization
  and mapping: Toward the robust-perception age,'' {\em IEEE Transactions on
  Robotics (T-RO)}, vol.~32, no.~6, pp.~1309--1332, 2016.

\bibitem{fuentes2015visual}
J.~Fuentes-Pacheco, J.~Ruiz-Ascencio, and J.~M. Rend{\'o}n-Mancha, ``Visual
  simultaneous localization and mapping: A survey,'' {\em Artificial
  Intelligence Review}, vol.~43, no.~1, pp.~55--81, 2015.

\bibitem{klein2007parallel}
G.~Klein and D.~Murray, ``Parallel tracking and mapping for small {AR}
  workspaces,'' in {\em International Symposium on Mixed and Augmented Reality
  (ISMAR)}, pp.~225--234, 2007.

\bibitem{mur2015orb}
R.~Mur-Artal, J.~M.~M. Montiel, and J.~D. Tardos, ``{ORB-SLAM}: A versatile and
  accurate monocular {SLAM} system,'' {\em IEEE Transactions on Robotics
  (T-RO)}, vol.~31, no.~5, pp.~1147--1163, 2015.

\bibitem{von2020gn}
L.~von Stumberg, P.~Wenzel, Q.~Khan, and D.~Cremers, ``{GN-Net}: The
  gauss-newton loss for multi-weather relocalization,'' {\em IEEE Robotics and
  Automation Letters (RA-L)}, vol.~5, no.~2, pp.~890--897, 2020.

\bibitem{gao2018ldso}
X.~Gao, R.~Wang, N.~Demmel, and D.~Cremers, ``{LDSO}: Direct sparse odometry
  with loop closure,'' in {\em IEEE/RSJ International Conference on Intelligent
  Robots and Systems (IROS)}, pp.~2198--2204, 2018.

\bibitem{engel2017direct}
J.~Engel, V.~Koltun, and D.~Cremers, ``Direct sparse odometry,'' {\em IEEE
  Transactions on Pattern Analysis and Machine Intelligence (TPAMI)}, vol.~40,
  no.~3, pp.~611--625, 2017.

\bibitem{yang2020d3vo}
N.~Yang, L.~v. Stumberg, R.~Wang, and D.~Cremers, ``{D3VO}: Deep depth, deep
  pose and deep uncertainty for monocular visual odometry,'' in {\em IEEE/CVF
  Conference on Computer Vision and Pattern Recognition (CVPR)},
  pp.~1281--1292, 2020.

\bibitem{rublee2011orb}
E.~Rublee, V.~Rabaud, K.~Konolige, and G.~Bradski, ``{ORB}: An efficient
  alternative to {SIFT} or {SURF},'' in {\em IEEE International Conference on
  Computer Vision (ICCV)}, pp.~2564--2571, 2011.

\bibitem{davison2007monoslam}
A.~J. Davison, I.~D. Reid, N.~D. Molton, and O.~Stasse, ``{MonoSLAM}: Real-time
  single camera {SLAM},'' {\em IEEE Transactions on Pattern Analysis and
  Machine Intelligence (TPAMI)}, vol.~29, no.~6, pp.~1052--1067, 2007.

\bibitem{jin2003semi}
H.~Jin, P.~Favaro, and S.~Soatto, ``A semi-direct approach to structure from
  motion,'' {\em The Visual Computer}, vol.~19, no.~6, pp.~377--394, 2003.

\bibitem{engel2014lsd}
J.~Engel, T.~Sch{\"o}ps, and D.~Cremers, ``{LSD-SLAM}: Large-scale direct
  monocular {SLAM},'' in {\em European Conference on Computer Vision (ECCV)},
  pp.~834--849, 2014.

\bibitem{agarwal2011building}
S.~Agarwal, Y.~Furukawa, N.~Snavely, I.~Simon, B.~Curless, S.~M. Seitz, and
  R.~Szeliski, ``Building rome in a day,'' {\em Communications of the ACM},
  vol.~54, no.~10, pp.~105--112, 2011.

\bibitem{sattler2011fast}
T.~Sattler, B.~Leibe, and L.~Kobbelt, ``Fast image-based localization using
  direct {2D}-to-{3D} matching,'' in {\em IEEE International Conference on
  Computer Vision (ICCV)}, pp.~667--674, 2011.

\bibitem{schonberger2016vote}
J.~L. Sch{\"o}nberger, T.~Price, T.~Sattler, J.-M. Frahm, and M.~Pollefeys, ``A
  vote-and-verify strategy for fast spatial verification in image retrieval,''
  in {\em Asian Conference on Computer Vision (ACCV)}, pp.~321--337, Springer,
  2016.

\bibitem{lowe2004distinctive}
D.~G. Lowe, ``Distinctive image features from scale-invariant keypoints,'' {\em
  International Journal of Computer Vision (IJCV)}, vol.~60, no.~2,
  pp.~91--110, 2004.

\bibitem{rosten2006machine}
E.~Rosten and T.~Drummond, ``Machine learning for high-speed corner
  detection,'' in {\em European Conference on Computer Vision (ECCV)},
  pp.~430--443, 2006.

\bibitem{calonder2010brief}
M.~Calonder, V.~Lepetit, C.~Strecha, and P.~Fua, ``{BRIEF}: Binary robust
  independent elementary features,'' in {\em European Conference on Computer
  Vision (ECCV)}, pp.~778--792, 2010.

\bibitem{schonberger2017comparative}
J.~L. Sch{\"o}nberger, H.~Hardmeier, T.~Sattler, and M.~Pollefeys,
  ``Comparative evaluation of hand-crafted and learned local features,'' in
  {\em IEEE/CVF Conference on Computer Vision and Pattern Recognition (CVPR)},
  pp.~1482--1491, 2017.

\bibitem{huang2017coarse}
S.~Huang, M.~Gong, and D.~Tao, ``A coarse-fine network for keypoint
  localization,'' in {\em IEEE International Conference on Computer Vision
  (ICCV)}, pp.~3028--3037, 2017.

\bibitem{suwajanakorn2018discovery}
S.~Suwajanakorn, N.~Snavely, J.~J. Tompson, and M.~Norouzi, ``Discovery of
  latent {3D} keypoints via end-to-end geometric reasoning,'' in {\em
  Conference on Neural Information Processing Systems (NeurIPS)},
  pp.~2059--2070, 2018.

\bibitem{balntas2016learning}
V.~Balntas, E.~Riba, D.~Ponsa, and K.~Mikolajczyk, ``Learning local feature
  descriptors with triplets and shallow convolutional neural networks.,'' in
  {\em British Machine Vision Conference (BMVC)}, 2016.

\bibitem{yi2016lift}
K.~M. Yi, E.~Trulls, V.~Lepetit, and P.~Fua, ``{LIFT}: Learned invariant
  feature transform,'' in {\em European Conference on Computer Vision (ECCV)},
  pp.~467--483, 2016.

\bibitem{revaud2019r2d2}
J.~Revaud, C.~De~Souza, M.~Humenberger, and P.~Weinzaepfel, ``{R2D2}: Reliable
  and repeatable detector and descriptor,'' in {\em Conference on Neural
  Information Processing Systems (NeurIPS)}, pp.~12405--12415, 2019.

\bibitem{detone2018superpoint}
D.~DeTone, T.~Malisiewicz, and A.~Rabinovich, ``{SuperPoint}: Self-supervised
  interest point detection and description,'' in {\em IEEE/CVF Conference on
  Computer Vision and Pattern Recognition Workshops}, pp.~224--236, 2018.

\bibitem{luo2020aslfeat}
Z.~Luo, L.~Zhou, X.~Bai, H.~Chen, J.~Zhang, Y.~Yao, S.~Li, T.~Fang, and
  L.~Quan, ``{ASLFeat}: Learning local features of accurate shape and
  localization,'' in {\em IEEE/CVF Conference on Computer Vision and Pattern
  Recognition (CVPR)}, pp.~6589--6598, 2020.

\bibitem{galvez2012BoW}
D.~G\'alvez-L\'opez and J.~D. Tard\'os, ``Bags of binary words for fast place
  recognition in image sequences,'' {\em IEEE Transactions on Robotics (T-RO)},
  vol.~28, pp.~1188--1197, October 2012.

\bibitem{grauman2005pyramid}
K.~Grauman and T.~Darrell, ``The pyramid match kernel: Discriminative
  classification with sets of image features,'' in {\em IEEE International
  Conference on Computer Vision (ICCV)}, vol.~2, pp.~1458--1465, 2005.

\bibitem{lazebnik2006beyond}
S.~Lazebnik, C.~Schmid, and J.~Ponce, ``Beyond bags of features: Spatial
  pyramid matching for recognizing natural scene categories,'' in {\em IEEE/CVF
  Conference on Computer Vision and Pattern Recognition (CVPR)}, vol.~2,
  pp.~2169--2178, 2006.

\bibitem{fischler1981random}
M.~A. Fischler and R.~C. Bolles, ``Random sample consensus: a paradigm for
  model fitting with applications to image analysis and automated
  cartography,'' {\em Communications of the ACM}, vol.~24, no.~6, pp.~381--395,
  1981.

\bibitem{grisetti2011g2o}
G.~Grisetti, R.~K{\"u}mmerle, H.~Strasdat, and K.~Konolige, ``g2o: A general
  framework for (hyper) graph optimization,'' in {\em IEEE International
  Conference on Robotics and Automation (ICRA)}, pp.~9--13, 2011.

\bibitem{wenzel20204seasons}
P.~Wenzel, R.~Wang, N.~Yang, Q.~Cheng, Q.~Khan, L.~von Stumberg, N.~Zeller, and
  D.~Cremers, ``{4Seasons}: A cross-season dataset for multi-weather {SLAM} in
  autonomous driving,'' in {\em German Conference on Pattern Recognition
  (GCPR)}, 2020.

\bibitem{maddern2017Oxford}
W.~Maddern, G.~Pascoe, C.~Linegar, and P.~Newman, ``1 year, 1000 km: The
  {Oxford} {RobotCar} dataset,'' {\em International Journal of Robotics
  Research (IJRR)}, vol.~36, no.~1, pp.~3--15, 2017.

\bibitem{maddern2020real}
W.~Maddern, G.~Pascoe, M.~Gadd, D.~Barnes, B.~Yeomans, and P.~Newman,
  ``Real-time kinematic ground truth for the {Oxford} {RobotCar} dataset,''
  {\em arXiv preprint arXiv:2002.10152}, 2020.

\bibitem{sturm2012benchmark}
J.~Sturm, N.~Engelhard, F.~Endres, W.~Burgard, and D.~Cremers, ``A benchmark
  for the evaluation of {RGB-D} {SLAM} systems,'' in {\em IEEE/RSJ
  International Conference on Intelligent Robots and Systems (IROS)},
  pp.~573--580, 2012.

\end{thebibliography}
